# Exploring the Power of Topic Modeling Techniques in Analyzing Customer Reviews: A Comparative Analysis

Anusuya Krishnan* and Kennedyraj

**Abstract:** The exponential growth of online social network platforms and applications has led to a staggering volume of user-generated textual content, including comments and reviews. Consequently, users often face difficulties in extracting valuable insights or relevant information from such content. To address this challenge, machine learning and natural language processing algorithms have been deployed to analyze the vast amount of textual data available online. In recent years, topic modeling techniques have gained significant popularity in this domain. In this study, we comprehensively examine and compare five frequently used topic modeling methods specifically applied to customer reviews. The methods under investigation are latent semantic analysis (LSA), latent Dirichlet allocation (LDA), non-negative matrix factorization (NMF), pachinko allocation model (PAM), Top2Vec, and BERTopic. By practically demonstrating their benefits in detecting important topics, we aim to highlight their efficacy in real-world scenarios. To evaluate the performance of these topic modeling methods, we carefully select two textual datasets. The evaluation is based on standard statistical evaluation metrics such as topic coherence score. Our findings reveal that BERTopic consistently yield more meaningful extracted topics and achieve favorable results.

## 1 Introduction

In today's digital age, customer reviews have become an indispensable source of information for consumers and businesses alike. With the proliferation of e-commerce platforms, social media, and online review websites, customers have the power to share their opinions about products or services.

- Anusuya Krishnan is with the Data science research group in College of information technology, United Arab Emirates University, UAE . E-mail: anusuyababy18@gmail.com
- Kennedyraj is with the AI group, Noorul Islam University, Kanyakumari, India. E-mail:kennedycivil@gmail.com

* To whom correspondence should be addressed.

These customer reviews provide valuable insights into the strengths, weaknesses, and overall quality of various offerings, helping other potential customers make informed decisions. However, the sheer volume of customer reviews available online presents a challenge. Businesses often struggle to manually analyze and extract meaningful information from this vast amount of unstructured data. This is where topic modelling comes into play[1].

Topic modelling is a powerful computational technique that enables businesses to uncover underlying themes, topics, and patterns within a collection of customer reviews. By applying natural language processing (NLP) and machine learning algorithms, topic modelling algorithms can automatically identify and extract key topics from textual data. This allows businesses to gain a deeper understanding of customer opinions, preferences, and concerns.

Topic modeling can be used with customer reviews to gain insights into what customers are saying about your products or services. For example, one could use topic modeling to identify the different features that customers are most satisfied with, or the areas where one need to improve their own products or services. Topic modeling can also be used to identify customer pain points. For example, if the online food platform sees that a lot of customers are talking about the same problem in their reviews, they can use this information to improve their product or service.

The application of topic modelling to customer reviews offers several benefits. Firstly, it provides a systematic and efficient way to analyze large volumes of customer feedback. Instead of manually reading and categorizing each review, businesses can leverage topic modelling algorithms to automate the process. This not only saves time and resources but also enables businesses to extract insights on a scale that would be otherwise impossible[2].

Secondly, topic modelling allows businesses to identify emerging trends and recurring themes within customer reviews. By uncovering the most frequently discussed topics, businesses can gain valuable insights into what aspects of their products or services are resonating with customers and what areas may require improvement. This information can inform decision-making processes, such as product development, marketing strategies, and customer service enhancements.

Furthermore, topic modelling can help businesses monitor and assess sentiment within customer reviews. By associating sentiment scores with different topics, businesses can identify which aspects of their offerings are receiving positive or negative feedback. This knowledge can be leveraged to prioritize areas of improvement or highlight positive features in marketing campaigns[3].

Over the years, extensive research has been conducted in this field, leading to the development of various methods. These methods can be broadly classified into three main categories: algebraic, probabilistic, and neural models[4-5]. The algebraic and probabilistic models are traditional statistical approaches, while the neural model represents the latest advancement in this area, leveraging artificial neural networks and benefiting from the widespread adoption of deep learning in NLP.

Algebraic models encompass a range of techniques such as latent semantic indexing (LSI) and non-negative matrix factorization (NMF)[25-26]. Probabilistic models include probabilistic latent semantic indexing (pLSI), anchored correlation explanation (CorEx), latent Dirichlet allocation (LDA), and various extensions and variants of LDA like hierarchical Dirichlet process (HDP), correlated topic model (CTM), and structural topic model (STM) [6-8]. LDA has been the most widely used method for decades, but its adoption in research studies has often been unquestioned, primarily due to its popularity, without providing a rationale for selecting the topic modeling method[9]. LDA can be seen as a generalized version of pLSI, incorporating a Dirichlet prior distribution across document-topic and topic-word distributions. One limitation of LDA is its reliance on the bag-of-words (BoW) representation, which disregards the semantic relationships between words within a text[9].

More recently, neural models have gained significant popularity since 2016, coinciding with the rapid advancements in deep learning. Examples of neural models include lda2vec, SBM, deep LDA, Top2Vec, and BERTopic[10-14, 26-28]. The development of these models aligns with the exponential growth of deep learning techniques. For instance, deep LDA is a hybrid model that combines LDA with a basic multilayer perceptron (MLP) neural network. In contrast, the more recent BERTopic utilizes bidirectional encoder representations from transformers (BERT) and class-based TF-IDF (c-TF-IDF) to achieve state-of-the-art performance. BERTopic has emerged as a dominant approach in the topic modeling field, showcasing impressive results on various datasets with minimal data preprocessing requirements[14].

In this study, our primary aim is to evaluate and compare the performance of five topic modeling techniques: LDA, NMF, LSA, PAM, and BERTopic. These techniques provide unique methodologies for extracting topics from textual data. Latent Dirichlet Allocation (LDA) is a generative statistical model that identifies topics based on the probability distribution of words[8]. Non-Negative Matrix Factorization (NMF) utilizes a linear algebra approach to uncover latent topics by factorizing the term-document matrix[9]. Latent Semantic Analysis (LSA) employs a dimensionality reduction technique to capture semantic relationships between words and documents[10]. Pachinko Allocation Model (PAM) integrates topic modeling with authorship attribution to uncover both topic and author information[11-12]. Top2Vec combines the strengths of word2vec, clustering, and document embeddings to identify coherent topics in text data[13]. Lastly, BERTopic leverages the power of BERT (Bidirectional Encoder Representations from Transformers) embeddings to identify coherent topics in text[14]. By examining and comparing the effectiveness of these techniques, we aim to gain insights into their respective strengths and weaknesses in topic modeling applications. To evaluate the performance of the five different models, a coherence score was calculated.

The organization of this paper is as follows: In Section 2, we provide a concise literature review of previous studies that have focused on extracting topics using machine learning and natural language processing techniques. Section 3 presents the framework proposed in this study. The obtained results are reported in Section 4, and the future implications of our work are discussed in Section 5.

The font is Times New Roman. The text may include an introduction, experimental details, theoretical basis, results, discussion, and conclusions. For articles, please use the decimal system of headings with no more than three levels.

# 2 Related work

In recent years, there has been a notable increase in research dedicated to the field of topic modeling for customer reviews. Scholars have extensively explored different techniques, models, and applications, shedding light on the advantages and hurdles associated with employing topic modeling for analyzing customer feedback. This growing attention stems from the necessity to efficiently analyze vast amounts of text data and unveil underlying themes and patterns within it. Researchers have ventured into diverse methodologies, merging the capabilities of machine learning with topic modeling techniques to enhance their analyses [1]. In the last ten years, topic modeling (TM) has been extensively applied across diverse domains, including health, hospitality, education, social networks, and finance[15]. Its adoption has proven beneficial for both academic and industrial purposes, particularly in interdisciplinary studies. However, recent literature indicates a trend where most studies concentrate on the application of a specific TM technique, often limited to one technique, within a particular domain.

In a study conducted by researchers, a corpus of webpages related to food safety issues was analyzed using Latent Dirichlet Allocation (LDA)[16]. The study demonstrated the potential of LDA as a valuable tool for communication researchers in identifying websites relevant to food safety. Also, authors conducted a smart literature review by applying LDA to research papers[17]. By generating topics using LDA, the authors were able to select appropriate topics for their literature review, facilitating an efficient and focused review process. Another study focused on hotel reviews in New York City[18]. The authors employed Structural Topic Modeling (STM) to analyze the reviews and highlighted the improvement in inferences about consumer dissatisfaction through the use of STM.

A research paper utilized STM to identify research topics from the title, keywords, and abstract of articles published in the journal Computers & Education over a span of forty two years[19]. This approach allowed for the identification of recurring themes and trends within the field. Researchers investigated COVID-related news from outlets in the UK, India, Japan, and South Korea[20]. They also employed the Top2Vec algorithm to identify widely reported topics and subsequently performed sentiment analysis using the RoBERTa model. A study focused on analyzing bitcoin-related posts on Twitter, Reddit, and Bitcoin Talk. They investigated LDA to extract topics from the posts, which were then utilized by an LSTM-based neural system for stock price prediction[21].

Another study adopted and compared four topic modeling techniques, including LDA, NMF, Top2Vec, and BERTopic, on social media data for social science research. NMF and BERTopic demonstrated better performance compared to the other two techniques in this scenario[2]. A research paper analyzed tweets related to the COVID-19 vaccine. LDA was employed to extract topics from the tweets, and a sentiment polarity analysis using a dictionary-based method was performed[22].

In another study finding, three topic modeling techniques (LDA, CorEx, and NMF) were employed to identify traveler experiences from Instagram posts using a specific hashtag[23]. Lastly, some of the authors applied LDA to analyze financial news. The study aimed to highlight predictions and speculative statements within the news articles through a graphical user interface[24].

LDA has emerged as the predominant topic modeling (TM) technique in recent literature, as evidenced by studies[17-23]. These studies collectively showcase the potential of TM in uncovering topics across diverse fields. However, there is a scarcity of research utilizing TM in online review-related tasks. A recent study, for instance, employed LDA to identify topics from short tweets and generate predictions specifically for financial instruments[24]. These studies demonstrate the diverse applications of topic modeling in various domains and the effectiveness of different algorithms in extracting valuable insights from different types of textual data.

The comprehensive review study explores different models and frameworks for opinion mining and sentiment analysis, including topic modelling. It discusses the role of topic modelling in identifying and extracting key topics from customer reviews and emphasizes its usefulness in understanding customer sentiments and preferences. Researchers focus on the application of topic modelling in sentiment analysis of customer reviews. The authors propose a novel approach that combines topic modelling with sentiment analysis to uncover hidden topics and sentiments within customer feedback[28]. The study demonstrates the effectiveness of this approach in identifying aspects of products or services that contribute to positive or negative sentiments[29].

Another study explores the use of topic modelling to discover user interests in online reviews. The authors propose a topic-based interest model that captures user preferences and interests based on the topics extracted from customer reviews. The findings demonstrate the effectiveness of topic modelling in understanding user preferences and improving personalized recommendation systems[30].

This study focuses on the application of topic modelling for sentiment analysis in the hospitality industry. The authors explore the use of Latent Dirichlet Allocation (LDA) to extract topics from customer reviews in the hotel domain. The findings highlight the benefits of topic modelling in identifying specific areas of concern or satisfaction within customer feedback[31].

## 3 Methodology

The present study examines a selection of topic modeling (TM) methods through a comprehensive review. Specifically, we focus on six commonly employed TM techniques that utilize diverse representation forms and statistical models. Figure 1 illustrates a standard process for topic generation, which serves as the foundation for our analysis.

Our evaluation encompasses both topic quality and performance metrics. It is important to note that the fundamental differences among these methods lie in their approaches to capturing structures and the specific aspects of those structures they exploit. While the field of social media textual data encompasses numerous TM methods, we have selected the most popular ones for the purpose of comparison, as it is impractical to mention every single method.

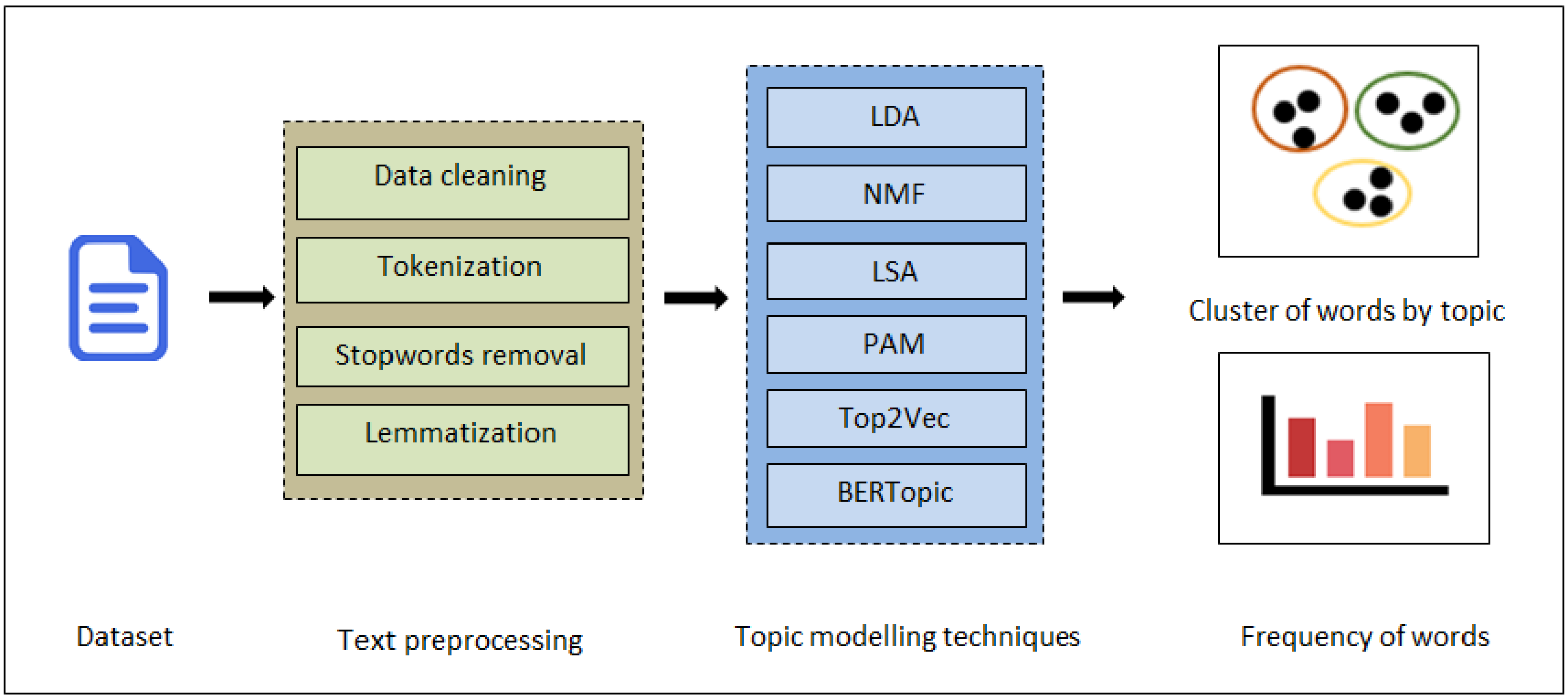

**Fig. 1 Overview of topic modelling schema**

## 3.1 Data Collection

In this study, two datasets were utilized: the OpSpam dataset and the Customer Satisfaction dataset from the UAE Ministry of Economy (MOE) government website.

The first dataset comprises 29,200 reviews obtained from the UAE Ministry of Economy. These reviews are authentic and reflect the customer satisfaction levels with the MOE app. They provide insights into the experiences and opinions of real users regarding the services offered by the government.

The second dataset is a OpSpam dataset which is publicly available in kaggle. It comprises 1600 customer reviews[32]. These reviews were collected from popular platforms such as TripAdvisor and Amazon Mechanical Turk. The dataset serves as a valuable resource for sentiment analysis and opinion mining tasks, providing a diverse range of reviews for analysis and research purposes. Researchers and practitioners can leverage this dataset to study customer opinions, sentiment trends, and related topics in various domains.

The second dataset comprises 29,200 reviews obtained from the UAE Ministry of Economy. These reviews are authentic and reflect the customer satisfaction levels with the MOE app. They provide insights into the experiences and opinions of real users regarding the services offered by the government.

## 3.2 Data preprocessing

The data preprocessing module plays a crucial role in refining and cleaning the raw data, ensuring that it contains only the necessary features for the topic modelling task. This module employs various techniques to enhance the quality of the data. It effectively removes irrelevant phrases and symbols, eliminating any noise or distractions from the text. Additionally, stopword elimination is applied to remove commonly used words in the language that do not contribute significantly to text mining tasks. These words, such as prepositions, numbers, and other irrelevant terms, lack relevant information for the study.

To facilitate deep analysis, tokenization is employed, dividing the text input into meaningful units such as phrases, words, or tokens. The outcome of tokenization is a sequence of these tokens, which serve as the fundamental units for further analysis and processing. Furthermore, lemmatization with part-of-speech (POS) tagging is applied to optimize the data. This process reduces the feature space by mapping words to their base or dictionary forms, reducing redundancy and improving the efficiency of subsequent analyses.

## 3.3 Topic modeling techniques

Topic modeling is a widely used technique in text mining that aims to identify clusters of words, known as topics, that frequently appear together in a corpus of documents. These topics are discovered using probabilistic models. Essentially, a topic represents a probability distribution over all the words in the vocabulary of the corpus, where the most likely words are indicative of its content. Topic modeling relies on numerical vectors, such as document × term vectors, to process the input documents. These vectors are then transformed into topic × term and document × topic vectors. The assignment of words to specific topics is accomplished through the document × term matrix. In this study, various topic modeling methodologies are employed, each with its own approach. These methodologies will be briefly described in the following sections.

### 3.3.1 LDA

Latent Dirichlet Allocation (LDA) is a widely recognized generative probabilistic model utilized to uncover latent topic information within a large collection or corpus of documents[8]. This model employs the bag-of-words (BoW) approach, treating each document as a vector of word frequencies. By doing so, it transforms textual information into numerical data that can be easily analyzed. LDA effectively reduces the dimensionality of the BoW model by representing a document as a combination of topics[12]. Typically, a few hundred topics are used, resulting in a vector representation of the document with a few hundred dimensions. This dimension reduction greatly accelerates training and minimizes the risk of overfitting. Mathematically, the LDA model can be summarized as follows:

$$\Phi_n \sim \text{Dirichlet}(\beta) \tag{1}$$

$$\Theta_p \sim \text{Dirichlet}(\alpha) \tag{2}$$

In the above formulation, Eq. (1) represents topic word distribution and equation Eq. (2) represents document topic distribution. $\Phi_n$ denotes the distribution of words for topic n, and $\Theta_p$ denotes the distribution of topics for document p. The model hyperparameters α and β represent the document-topic density and topic-word density respectively.

### 3.3.2 LSA

The Latent Semantic Analysis (LSA) method adopts a two-step approach for topic learning[10]. In the first step, it constructs a conventional term-by-document matrix, a common technique in information retrieval. To enhance the significance of informative words, the matrix counts are then smoothed. The log-entropy transform is applied to improve upon the original LSA model. In the second step, LSA decomposes the smoothed matrix, revealing general patterns and connections between words and documents. LSA efficiently handles large volumes of raw text by breaking them down into individual words and organizing them into coherent sentences or paragraphs. It considers both the similarity between terms within a text and their relationships to other terms, leading to a deeper understanding of the underlying topic structure. Moreover, the LSA model generates vector-based representations for texts, facilitating the grouping of related words. The LSA method learns latent topics through matrix decomposition of the term-document matrix[10]. Let's assume we have a term-by-document matrix X, which can be decomposed into three other matrices: a, b, and c. By multiplying these matrices together, we obtain the reconstructed matrix y, expressed in Eq. (3) as:

$$\{y\} = \{a\}\{b\}\{c\} \tag{3}$$

In this decomposition, the columns represent paragraphs, while the rows represent unique words..

### 3.3.3 NMF

Non-negative Matrix Factorization (NMF) is another approach to factorize matrix M by minimizing the reconstruction error, but with the additional constraint that the decomposed matrices contain only non-negative values[9]. In this sense, NMF can be seen as learning unnormalized probability distributions over topics. Formally, NMF is defined as:

$$V = FB \tag{4}$$

In Eq. (4), the term-document matrix V of size (m×n) represents the relationship between m terms and n documents. The matrix F, with dimensions k×n, represents the topic distribution over the documents, where k is the number of topics. Each column of F corresponds to a topic and its associated term weights. Additionally, the matrix B, of size (k×n), captures the document weights for each topic. Each column of B corresponds to the document weights for the respective topic. By enforcing non-negativity, NMF provides a useful framework for exploring and interpreting topics in various applications.

### 3.3.4 PAM

The Pachinko Allocation Model (PAM) is an advanced topic modeling technique that addresses some of the limitations of the Latent Dirichlet Allocation (LDA) model. PAM is designed to capture the hierarchical structure of topics within a document collection. PAM introduces a hierarchical structure to the topic modeling process. It represents topics as a directed acyclic graph (DAG) or a tree structure. This hierarchy allows for the modeling of relationships between topics at different levels.

In PAM, each document is associated with a set of topics that are selected from different levels of the hierarchy. The model considers both the document-specific topics and the higher-level topics that provide context and coherence to the document's theme. This hierarchical organization enables PAM to capture more complex topic relationships and provide more accurate representations of documents. The probability of generating a whole corpus, $P(E|\alpha)$, in the context of topic modeling can be calculated as the product of the probability for each individual document, $P(n|\alpha)$, where $\alpha$ represents the hyperparameters of the model. Mathematically, PAM can be expressed in Eq. (5) as:

$$P(E|\alpha) = \Pi \, P(n|\alpha) \tag{5}$$

Here, $P(n|\alpha)$ represents the probability of generating an individual document ‘n’ given the hyperparameters $\alpha$. The product $\Pi$ denotes the multiplication of these probabilities for all documents in the corpus.

### 3.3.5 Top2Vec

Top2Vec is an innovative unsupervised machine learning approach that offers scalable and effective solutions for topic modeling and document clustering. Its primary objective is to identify relevant themes within large-scale text corpora[13].Top2Vec maps both documents and words to a shared semantic vector space using the Doc2Vec method. The document vectors are subsequently clustered, resulting in multiple clusters, each representing a distinct topic. The topic representation of a cluster is determined by averaging the document vectors within that cluster and extracting the nearest N words to the topic vector. One notable feature of Top2Vec is its ability to handle multi-word phrases and infrequently used terms, distinguishing it from traditional topic modeling techniques.

The Top2Vec process involves several steps. First, embedding vectors and words are generated using methods like Doc2Vec. Next, the dimensionality of the embedding vectors is reduced, often through techniques like UMAP. Subsequently, clustering is performed on the reduced vectors, typically using HDBSCAN. The centroids of the resulting clusters are then computed, representing distinct topics. Each topic's vector is derived by averaging the document vectors within its corresponding cluster. Finally, words closely associated with each cluster's vector are assigned to the respective topic.

Overall, Top2Vec offers a novel approach to scalable topic modeling and document clustering. By leveraging word embedding semantic similarity and hierarchical clustering, it provides an effective solution for extracting meaningful topics from large text corpora. Its ability to handle multi-word phrases and infrequent terms further enhances its capabilities, setting it apart from traditional topic modeling methods.

### 3.3.6 BERTopic

BERTopic is an advanced pre-trained topic modeling technique that utilizes BERT and c-TF-IDF to create dense clusters, enabling the interpretation of topics while preserving important words in topic descriptions[14]. Unlike traditional topic modeling approaches, BERTopic harnesses the contextualized word embeddings provided by BERT, allowing it to capture the semantic meaning and context of words within a corpus. Additionally, BERTopic provides a user-friendly interface that enables researchers to observe and analyze the results of the topic modeling process.

Similar to Top2Vec, BERTopic involves several steps. It begins by embedding documents, followed by dimensionality reduction using techniques like UMAP. Next, clustering is performed using methods such as HDBSCAN, resulting in the formation of clusters. From these clusters, topic representations are generated. However, what sets BERTopic apart is the incorporation of c-TF-IDF in the final step. This technique is employed to extract topic words, reduce the number of topics, and enhance word coherence and diversity using maximum marginal relevance (MMR).

### 3.4 Evaluation metric

Evaluating topic models does not have a one-size-fits-all approach. Coherence scores are considered the most appropriate metric when the output of topic modeling is used by human users, as stated in the study[7]. Different coherence scores, such as c_v and u_mass, are available for evaluation. Both (c_v and u_mass) coherence metrics assess the coherence of a topic by calculating the sum of pairwise distributional similarity scores among the words in the topic set, TS[34]. In general, this can be expressed in Eq. (6) as:

$$\text{Coherence (TS)} = \sum(w_i, w_j, \varepsilon) \quad (6)$$

Here, TS represents the set of words describing the topic ($w_i$, $w_j$), and ε denotes a smoothing factor that ensures the coherence score returns real numbers. The c_v score constructs content vectors based on word co-occurrences and calculates the score using cosine similarity and pointwise mutual information (PMI). On the other hand, the u_mass metric focuses on analyzing the word distribution within a topic to determine its coherence. Generally, higher coherence scores indicate better topics. However, it is important to note that coherence alone may not provide a perfect measure and human inspection of the results for interpretability is often necessary to complement coherence scores. To consider efficiency, especially when working with large document sets, we also evaluated the computation time or training time of each model as an indicator of efficiency in this experiment.

## 4 Experimental Results

In this section, we present the experimental setup and findings of our study, focusing on comparing the outcomes of different topic modeling (TM) approaches. The goal of these experiments is to provide a comprehensive analysis of the results obtained from each TM method in the given context. We delve into the details, highlighting the key findings and insights derived from our comparisons. By examining the outcomes of these experiments, we gain a deeper understanding of the performance and effectiveness of various TM techniques.

### 4.1 Experiment 1: Customer happiness dataset

In this section, we cleaned the raw dataset to structured format. To prepare the data, a cleaning process was implemented, involving the removal of redundant characters, numbers, stopwords, and symbols. Additionally, lemmatization was performed with POS tagging to ensure the extraction of meaningful sentences. In the next step, we applied topic meddling techniques to this cleaned data.

**Table 1 Performance Comparison of different topic modelling techniques in dataset 1**

| Topic modeling techniques | Coherence Score | |
|---|---|---|
| | K=5 | K = 10 |
| LDA | 0.45 | 0.40 |
| NMF | 0.49 | 0.50 |
| LSA | 0.50 | 0.53 |
| PAM | 0.49 | 0.44 |
| Top2Vec | 0.56 | 0.54 |
| BERTopic | 0.62 | 0.56 |

During the data extraction stage, our objective was to extract topics from the input data. We conducted multiple evaluations by varying the number of topics (t = 5, 10). The initial results of topic performance and coherence, analyzed using common standard metrics applicable to topic modeling (TM) methods, are presented in Table 1.

BERTopic demonstrated superior performance in terms of coherence and interpretability when applied to dataset 1. It outperformed all other topic models, achieving the highest coherence scores (c_v = 0.62, u_mass = -1.156). LSA had slightly lower coherence scores (c_v = 0.56) compared to Top2vec (c_v = 0.56). However, LDA yielded the lowest coherence score (c_v = 0.40). PAM and NMF performed moderately and achieved moderate coherence scores.

**Table 2 Sample topics generated by BERTopic (top 5 topics)**

| Topics | Top 5 keywords |
|---|---|
| 0 | Thanks, great, fast, approval, easy |
| 1 | Excellent, immediate, support, thing, pay |
| 2 | Good, bad, thank, application, time |
| 3 | Happy, thanks, happiness, user, satisfied |
| 4 | Excellent, service, job, keep, comment |

Table 2 showcases the top 5 keywords generated by BERTopic. Notably, BERTopic excelled in producing descriptive topics compared to other methods.

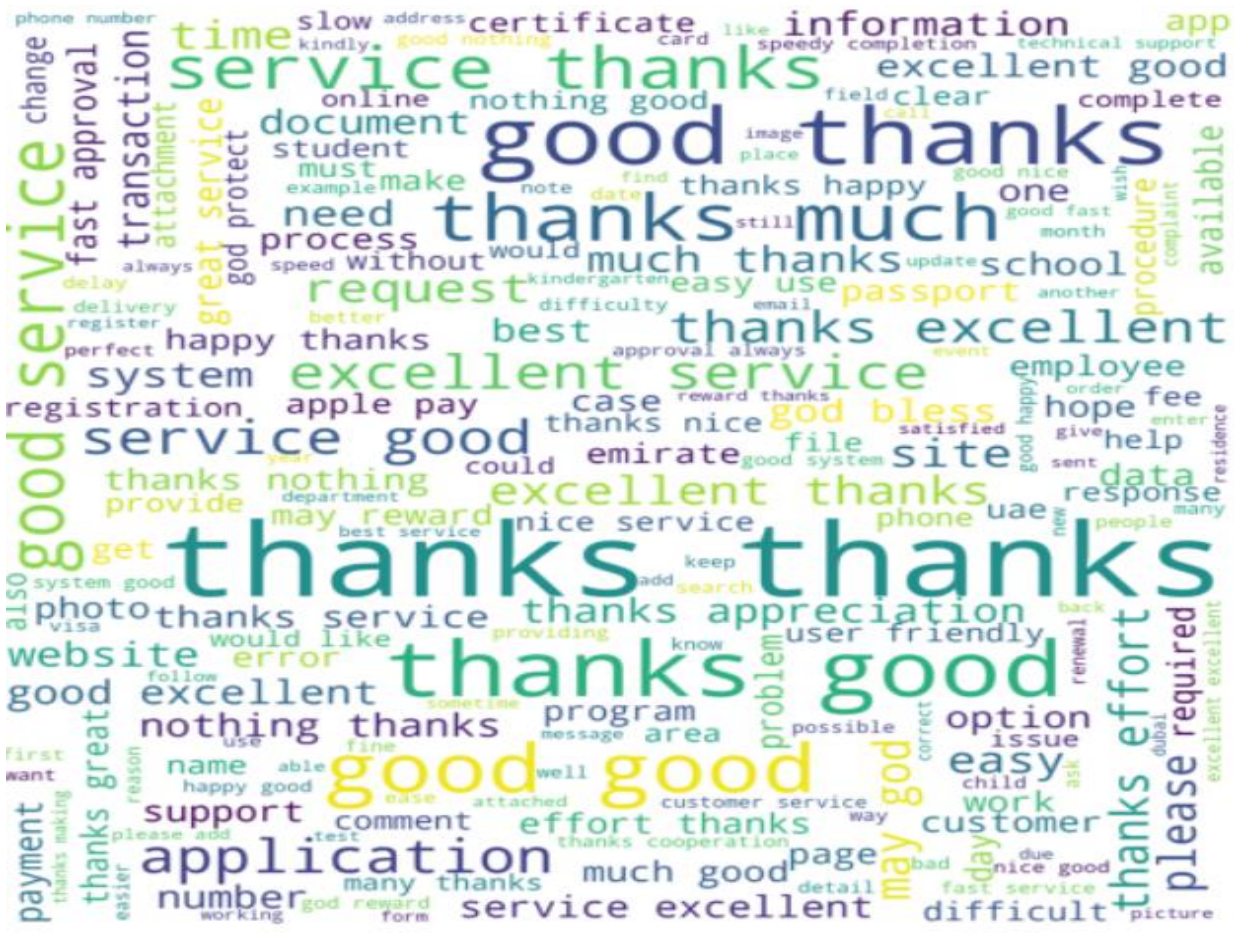

**Fig. 2 Word cloud for BERTopic-generated topics based on dataset 1**

Figures 2 showcases the word cloud graphs for BERTopic generated topics. The word cloud serves as a visual representation tool for topics, highlighting the most relevant words through the use of vivid colors and larger text. The word "thanks" and "good" are repeated and have strongest influence for detecting the customer review. To gain a comprehensive understanding of the significance of each term, generating a word cloud can offer a visually enhanced representation and shown in figure 2. This visualization technique aids in providing an overview of the importance and frequency of different terms within a given context.

## 4.2 Experiment 2: Hotel review dataset

For this experiment, we followed the same procedure as described in experiment 1. In dataset 2, BERTopic outperformed all other topic models, exhibiting the highest coherence scores (c_v = 0.60, u_mass = -2.532). LDA had slightly lower coherence scores (c_v = 0.42) compared to LSA (c_v = 0.46) for the top five topics. However, LDA displayed good coherence (c_v = 0.48) compared to NMF, which yielded the lowest coherence score (c_v = 0.33). PAM and Top2Vec performed poorly and achieved moderate coherence scores below 0.34.

**Table 3 Performance Comparison of different topic modelling techniques in dataset 2**

| Topic modeling techniques | Coherence Score | |
|---|---|---|
| | K=5 | K = 10 |
| LDA | 0.42 | 0.48 |
| NMF | 0.33 | 0.31 |
| LSA | 0.46 | 0.39 |
| PAM | 0.32 | 0.31 |
| Top2Vec | 0.38 | 0.33 |
| BERTopic | 0.58 | 0.60 |

Table 3 shows the performance comparison of different topic modeling techniques. As mentioned earlier, BERTopic performs well and achieve stable coherence score. Table 4 showcases a sample of topics generated by BERTopic, focusing on the top 5 topics and their corresponding keywords. The word "hotel" is repeated keyword in most of the topics.

**Table 4 Sample topics generated by BERTopic (top 5 topics)**

| Topics | Top 5 keywords |
|---|---|
| 0 | Hotel, great, room, day, staff |
| 1 | Get, immediate, night, day, bathroom |
| 2 | Good, floor, time, hotel, great |
| 3 | Stay, place, Chicago, hotel, front desk |
| 4 | Location, service, bathroom, problem, hotel |

These topics provide a glimpse into the key themes identified by BERTopic, showcasing the prominent keywords associated with each topic. During our analysis of the top 10 keywords in this experiment, we noticed that certain words such as "hotel," "stay," and "room" were frequently repeated and had a significant impact on detecting the hotel review topic. Figure 3 illustrates the word cloud generated by BERTopic visualization, further visualizing the prominence of these keywords within the topic.

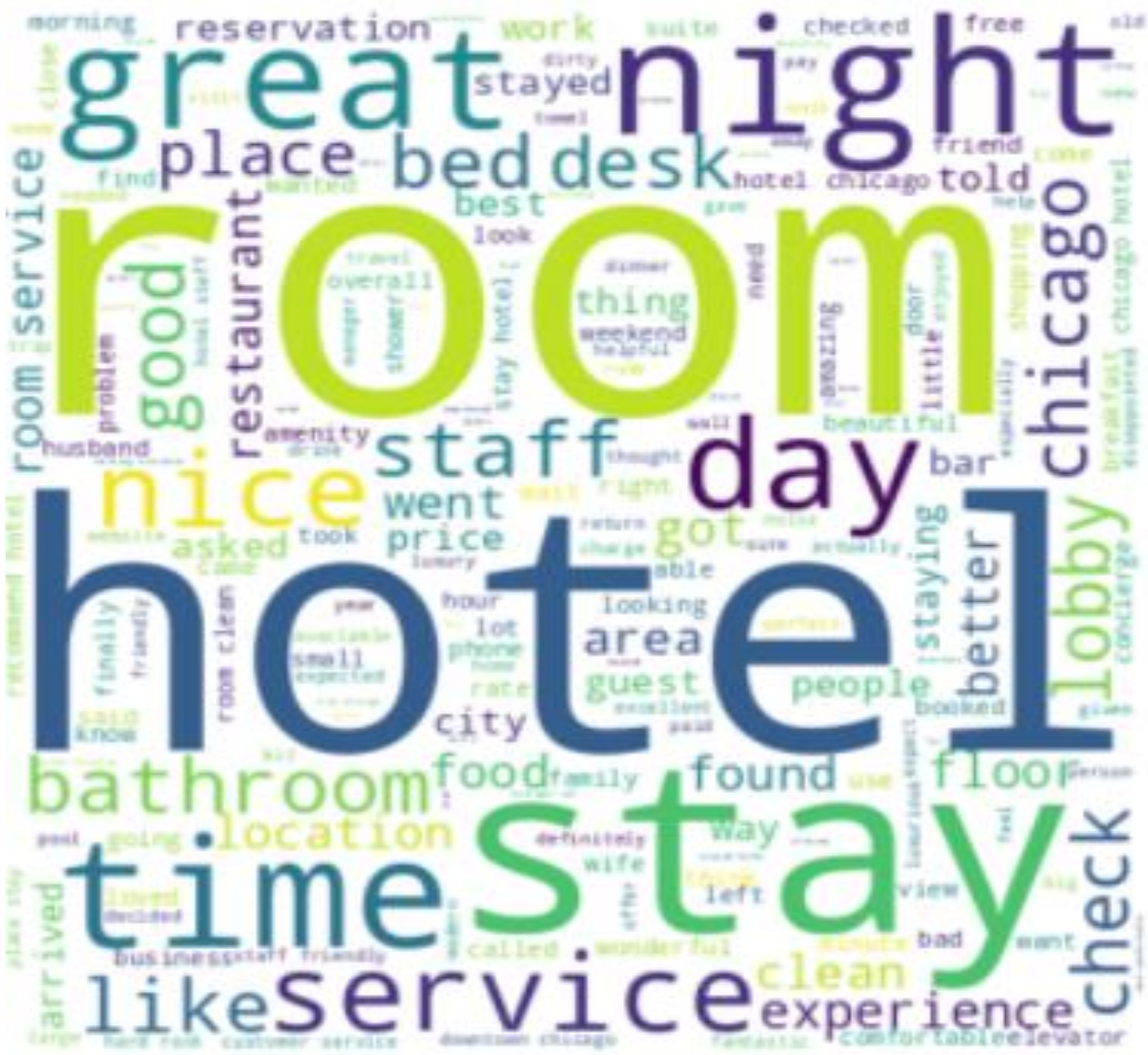

**Fig. 3 Word cloud for BERTopic-generated topics based on dataset 2**

Figure 3 illustrates the word cloud generated by BERTopic visualization, further visualizing the prominence of these keywords within the topic. Through this word cloud visualization, the most prominent and relevant words within each topic are visually highlighted. This enables a quick and intuitive understanding of the main themes and concepts captured by BERTopic.

## 5 Discussion

After evaluating our six models, it becomes evident that BERTopic outperforms the rest in terms of coherence scores. When considering dataset 1, which consists of short textual comments, LSA, NMF, and Top2Vec perform admirably, yielding coherence scores above 0.50. However, when dealing with dataset 2, comprising lengthier hotel review comments, the coherence score drops below 0.35. Nevertheless, BERTopic manages to maintain its superiority across both datasets, achieving an impressive coherence score of above 0.56.

From our observations in table 1, each TM method employed exhibited its own strengths and weaknesses. Throughout the evaluation, the results of all methods demonstrated similar performance. Overall, BERTopic yielded the highest term-topic probability, while Top2Vec, NMF, and LSA models displayed comparable performance. On the other hand, LDA coherence scores were comparatively lower than other methods in dataset 1. However in dataset 2, LDA performed well. NMF and PAM were lower than other models in dataset 2. In both datasets, LSA performed comparatively well. The term-topic probabilities ranged from 0 to 1 across all evaluated TM methods. However, it should be noted that certain conventional methods generated non-meaningful words, including domain-specific stop words, which were unsuitable for further processing.

Overall, BERTopic maintained consistent coherence for both the datasets across different parameter settings, suggesting that the parameters, such as the number of topics and maximum iterations, had limited impact on coherence.

In terms of computation efficiency, BERTopic surpassed Top2Vec in this scenario. Conventional methods like LDA, LSA, NMF, and PAM had computation times dependent on the number of topics, making them slower compared to BERTopic and Top2Vec for a similar number of topics (500). However, LDA could be faster with a smaller number of topics. It is essential to consider the trade-off between coherence and resource usage when selecting a topic model.

Regarding data preparation, conventional topic methods (LDA, LSA, NMF, PAM) required more complex preprocessing steps, including removing punctuation, useless symbols, stop words, and text normalization. In contrast, Top2Vec and BERTopic required minimal data preprocessing as their underlying models could comprehend the context with the original text structure. However, for small data samples, Top2Vec and BERTopic might generate topics with some stop words. This issue was mitigated in the news impact analysis scenario where the news documents were sufficiently long.

When it comes to parameter fine-tuning, both Top2Vec and BERTopic utilize HDBSCAN for clustering, which does not directly allow specifying the number of topics by default. However, BERTopic provides more customization options by exposing parameter settings of underlying components like UMAP and HDBSCAN. In contrast, conventional topic modeling methods (LDA, LSA, NMF, PAM) allow users to flexibly determine certain parameters, including the number of topics.

Taking all these factors into consideration, BERTopic emerged as the preferred choice for analyzing the impact of both long and short textual data on customer review analysis in this study. Its superior coherence, interpretability, and computational efficiency, along with the advantages of minimal data preprocessing and parameter customization, make it a reliable topic modeling approach.

## 6 Conclusion

The internet plays a crucial role in driving the demand for business applications and services that enhance shopping experiences and commercial activities for global customers. However, the vast amount of information and knowledge available online can sometimes overwhelm users, leading to additional time and effort spent in search of relevant information. The rise of online platforms like Twitter, Facebook, Instagram have emphasized the need for analyzing customer reviews, which pose challenges due to their limited and noisy nature, often resulting in inaccurate topic inference.

This research study presents a comparative analysis of six mainstream topic modeling techniques, namely LDA, LSA, PAM, NMF, and two emerging advanced neural model Top2Vec and BERTopic, in the context of customer review analysis on online social media platforms. The objective is to evaluate the performance of these topic models on short textual data and explore their integration into a customer review analysis scenario. The experimental results demonstrate that BERTopic emerges as the best-performing model overall, requiring minimal data preprocessing, achieving the highest coherence score, and demonstrating reasonable computing time.

However, this study acknowledges several limitations that open up potential avenues for future research. In this study, the six models (LDA, LSA, PAM, NMF, Top2Vec, and BERTopic) were compared in terms of coherence, interpretability, and computation time. Future research could involve comparing additional topic models. Furthermore, the evaluation of topic models lacks consensus regarding the appropriate measures to employ. Coherence and interpretability may not always be the most relevant evaluation metrics for different applications. In specific contexts, other measures such as topic diversity might take precedence. More research is necessary to study and analyze the strengths and weaknesses of various evaluation metrics. Furthermore, investigating the potential of large language models (LLMs) as an alternative to conventional topic models presents a promising direction, necessitating a comprehensive evaluation against established topic modeling techniques.